\documentclass[9pt]{article}

\usepackage{times}

\usepackage{soul}
\usepackage{url}
\usepackage[hidelinks]{hyperref}
\usepackage[utf8]{inputenc}
\usepackage[small]{caption}
\usepackage{graphicx}
\usepackage{amsmath}
\usepackage{booktabs}
\usepackage{geometry}
\usepackage{microtype}
\usepackage{subfigure}
\usepackage{booktabs} % for professional tables
\usepackage{multirow}
\usepackage{graphicx}
\usepackage{amsmath,amsthm}
\theoremstyle{definition}

\usepackage{xspace,color}
\usepackage{diagbox,multirow} 
\usepackage{subfigure}
\usepackage{tikz,bm}
\usepackage{algorithmic,algorithm}
\definecolor{PineGreen}{rgb}{0.0, 0.47, 0.44}
\usepackage{enumerate}
\usepackage{amsfonts}
\usepackage[utf8]{inputenc}
\usepackage[T1]{fontenc}    % use 8-bit T1 fonts
\usepackage{url}            % simple URL typesetting
\usepackage{booktabs}       % professional-quality tables 
\usepackage{nicefrac}       % compact symbols for 1/2, etc.
\usepackage{microtype}      % microtypography
\usepackage{tikz,bm}
\usepackage{adjustbox}      
\usepackage{filecontents}
\usepackage{array,rotating}
\usepackage[labelfont=bf]{caption}

\usepackage{caption}
\usepackage{xspace,color}
\usepackage{enumerate}
\usepackage{enumitem,amsmath,amssymb}
\usepackage{hyperref}
\newcommand{\EB}{\mathbb{E}}
\newcommand{\RB}{\mathbb{R}}

\newcommand{\bfv}{\mathbf{v}}
\newcommand{\bfx}{\mathbf{x}}

\newcommand{\bfS}{\mathbf{S}}
\newcommand{\bfG}{\mathbf{G}}
\newcommand{\bfg}{\mathbf{g}}
\newcommand{\bfD}{\mathbf{D}}
\newcommand{\bfI}{\mathbf{I}}
\newcommand{\bfO}{\mathbf{O}}
\newcommand{\bfz}{\mathbf{z}}

\newcommand{\sS}{\mathbf{S}}
\newcommand{\xx}{\mathbf{x}}

\newcommand{\bfQ}{\mathbf{Q}}

\newcommand{\calL}{\mathcal{L}}

\newcommand{\calF}{\mathcal{F}}
\newcommand{\calH}{\mathcal{H}}

\newcommand{\marked}{}
\newcommand{\mname}{\texttt{AntibodyFlow}\xspace}

\usepackage{amssymb,bbm}
\usepackage{amsmath,amsthm,amsfonts,booktabs}

\newtheorem{corollary}{Corollary}

\theoremstyle{definition}

\title{\mname: Normalizing Flow Model for Designing Antibody Complementarity-Determining Regions}

\author{Bohao Xu$^{1}$  Yanbo Wang$^{2}$ Wenyu Chen$^{2}$ and Shimin Shan$^{3}$}
\date{%
    $^1$ School of Electrical Engineering and Computer Science, Nanjing University of Aeronautics and Astronautics, NanJing\\
    $^2$ School of Computer Science and Technology, North University of China, Taiyuan\\
    $^3$ School of Semiconductor and Physics, North University of China, Taiyuan\\
}

\begin{document}

\maketitle

\begin{abstract}
Therapeutic antibodies have been extensively studied in drug discovery and development in the past decades. 
Antibodies are specialized protective proteins that bind to antigens in a lock-to-key manner. 
The binding strength/affinity between an antibody and a specific antigen is heavily determined by the complementarity-determining regions (CDRs) on the antibodies. 
Existing machine learning methods cast \textit{in silico} development of CDRs as either sequence or 3D graph (with a single chain) generation tasks and have achieved initial success.  
However, with CDR loops having specific geometry shapes, learning the 3D geometric structures of CDRs remains a challenge. 
To address this issue, we propose \mname, a 3D flow model to design antibody CDR loops. 
Specifically, \mname first constructs the distance matrix, then predicts amino acids conditioned on the distance matrix. 
Also, \mname conducts constraint learning and constrained generation to ensure valid 3D structures. 
Experimental results indicate that \mname outperforms the best baseline consistently with up to 16.0\% relative improvement in validity rate and 24.3\% relative reduction in geometric graph level error (root mean square deviation, RMSD). 
\end{abstract}

\section{Introduction}

% numerous ideal properties of antibodies including minimal adverse effects and the ability to bind to many ``undruggable'' targets due to different biochemical mechanisms~\cite{narayanan2021machine}. One critical task for antibody discovery, determining binding ability. 

Therapeutic antibodies have been extensively studied in drug discovery and development in the past decades~\cite{nelson2009development}. 
Compared to small-molecule drugs, therapeutic antibodies have many advantages such as minimal adverse effects, and can bind to many previously ``undruggable'' targets due to different biochemical mechanisms~\cite{narayanan2021machine}. 
% A natural human antibody has a symmetric Y shape; each half of the symmetric unit has two chains: a heavy chain (H) and a light chain (L). 
The majority of the affinity and specificity of antibodies is modulated by a set of binding loops called the Complementarity Determining Regions (CDRs) found on the variable domain of each of the two chains. 
One critical task for antibody discovery is to design the appropriate CDR~\cite{maccallum1996antibody,sela2013structural}. 
Traditionally, antibody sequences are discovered either \textit{in vivo} using animal immunization or by \textit{in vitro} affinity selection of candidates from large synthetic libraries of antibody sequences~\cite{kretzschmar2002antibody}. However, these approaches are time- and labor-intensive and unable to efficiently explore the large antibody space since the design process heavily relies on existing human knowledge~\cite{maccallum1996antibody}.

\textit{In silico} antibody CDR design methods were developed to tackle this challenge.  
Earlier methods are primarily based on heuristic searching and require extensive wet-lab validation~\cite{soderlind2000recombining,pantazes2010optcdr,liu2017computational}. 
In recent years, several sequence/3D graph generation methods have been proposed for antibody CDR loop discovery, 
e.g., ~\cite{akbar2021silico,saka2021antibody,jin2021iterative}. 
% genetic algorithm~\cite{jendrusch2021alphadesign}. 
However, several challenges remain: 
\begin{enumerate}[leftmargin=*]
\item 
Most of the existing methods generate amino acids sequentially, which is hard to model long-term dependency and impose global constraints on the geometric shape of CDR loops. 
\item 
Rotating and translating the 3D graph would not change the underlying 3D graph and it is challenging to directly generate rotation- and translation-invariant 3D graphs~\cite{satorras2021equivariant}. 
% and it is challenging to build rotation- translation-invariant 3D graph generator~\cite{satorras2021equivariant}. 
% \item To make sure the validity of the CDR loop, there are several geometric constraints on the geometry of the CDR loop. 
\end{enumerate}

To address these challenges, we propose \mname, a 3D flow model that generates antibody CDR loops in two phases: (i) build the geometric structure of the CDR loop with spatial constraints; and (ii) predict the amino acids as the graph nodes. 
The main contributions are summarized as follows.  
\begin{enumerate}[leftmargin=*]
% in a \textit{one-hot} manner (instead of sequential) so that we can impose global constraints. 
\item \mname generates amino acid sequence and pairwise distance matrix (between amino acids) in a \textit{one-hot} manner. 
Unlike the sequential manner, the flow model with one-hot generation is more flexible to impute the global constraints of geometric shape. 
Concretely, \mname designs constraint learning to learn the constraints and constrains 3D coordinates generation leveraging the constraints.  
% \item To guarantee the geometric validity (represented by constraints) of the generated CDR loop, we (1) encourage \mname to learn the validity constraints implicitly (Section~\ref{sec:constraint}) and (2) explicitly ensure the validity constraints when building a 3D graph (i.e., folded amino acid sequence).  
\item To circumvent the direct generation of 3D graphs, we build a pairwise distance matrix that measures Euclidean distances between amino acids. The distance matrix is invariant w.r.t. rotation and translation on a 3D graph. 
After that, we generate 3D coordinates based on both distance matrix and geometric constraints. 
\item Experiments confirmed \mname outperforms the best baseline consistently with up to 16.0\% relative improvement in validity rate and 24.3\% relative reduction in geometric graph level error (root mean square deviation, RMSD). 
\end{enumerate}

\section{Method}
\label{sec:method}

% Section~\ref{sec:background} formulates CDR design problem. Then Section~\ref{sec:flow} presents the background of the flow model and overview of our proposed flow model for antibody CDR generation \mname. The detailed methods are described further in two parts: distance matrix flow model (Section~\ref{sec:distance}) and amino acid flow model (Section~\ref{sec:amino}). Section~\ref{sec:constraint} discusses how the flow model learns constraints. Section~\ref{sec:coordinate_generation} describes how to generate 3D coordinates of CDR loops based on generated distance matrix and geometric constraints.  The learning and inference procedures are elaborated in Section~\ref{sec:learning}. 

\subsection{Problem Formulation}
\label{sec:background}

This section describes the antibody CDR loop and formulates the antibody CDR design problem. 
For ease of exposition, we list the important mathematical notations in Table~\ref{table:notation} in Appendix. 
% We slightly abuse the mathematical notations. 

\noindent\textbf{Amino acid}. A protein (including antibodies) is an amino acid sequence folded into a three-dimensional structure. 
Specifically, amino acids are small organic molecules that consist of an $\alpha$ (central) carbon atom linked to an amino group, a carboxyl group, a hydrogen atom, and a variable component called a side chain. 
Multiple amino acids are linked together within a protein by peptide bonds, thereby forming a long chain. 
Amino acids are the basic building blocks of a protein/antibody. 
The set of all the natural amino acids (denoted $V$) consists of 20 natural amino acids (i.e., $|V|=20$), e.g., Alanine, Cysteine, Glutamine, etc.

\begin{figure}
\centering
\includegraphics[width=0.74\columnwidth]{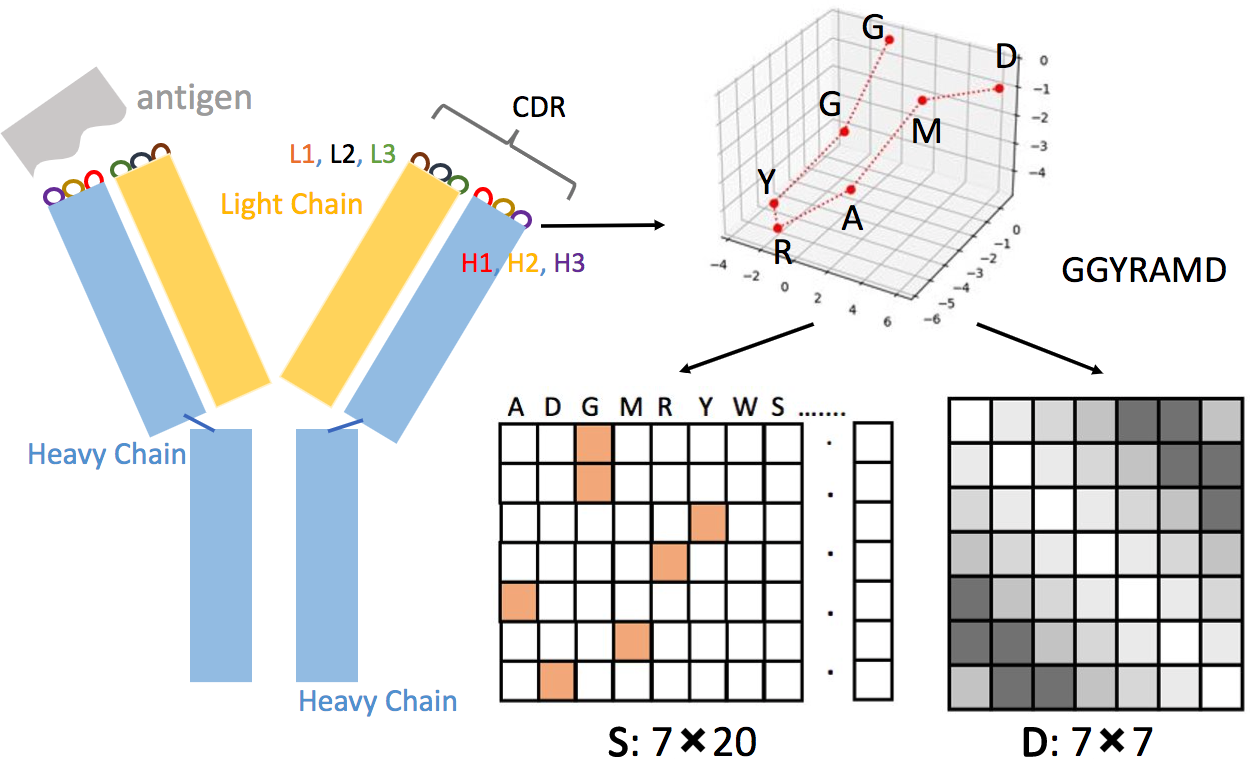}
\caption{Data representation. 
An antibody is a special kind of protein with a symmetric Y shape, each half of the symmetric unit has two chains: a heavy chain (H) and a light chain (L). In total, there are four chains, two identical H/L chains. 
The majority of the binding affinity (to specific antigens) is modulated by a set of binding loops called the Complementarity Determining Regions (CDRs) found on the variable domain of each of the H and L chains. 
There are 6 CDR loops on each half of the antibody, L1, L2, L3 on the light chain, and H1, H2, H3 on the heavy chain. 
\marked{We show the H3 loop of the antibody whose protein data bank (PDB) ID is ``5iwl'' as an example}. 
The CDR loop in the 3D geometry graph contains 7 amino acids (GGYRAMD) and their coordinates. 
The sequence is represented as a binary matrix $\bfS \in \{0,1\}^{7\times 20}$ (20 natural amino acids, each row is a one-hot vector), and the geometry information is represented as a pairwise distance matrix $\bfD\in\RB^{7\times 7}_{+}$. }
\label{fig:data}
\end{figure}

\noindent\textbf{CDR loops}. 
An antibody is a special kind of protein with a symmetric Y shape; each half of the symmetric unit has two chains: a heavy chain (H) and a light (L) chain, as shown in Figure~\ref{fig:data}.  
% The majority of the affinity and specificity of antibodies is modulated by a set of binding loops called the Complementarity Determining Regions (CDRs) found on the variable domain of each of the two chains. 
There are six CDR loops, L1, L2, L3 on the light chain and H1, H2, H3 on the heavy chain~\cite{maccallum1996antibody,sela2013structural}. 
CDR is complicated to model computationally, causing a significant hurdle for \textit{in silico} development of antibody biotherapeutics. 
Compared with L1, L2, and L3 loops, the H1, H2, and H3 loop in the CDR of an antibody plays a critical role in its binding ability to potential antigens~\cite{nelson2009development,regep2017h3}. Thus, this paper restricts our attention to modeling H1, H2, and H3 loops separately.

\noindent\textbf{Data representation}. 
A 3D CDR loop (H1, or H2, or H3) can be represented as $\sS$ amino acid sequence and their coordinates $\bfG)$. 
Given a CDR loop of  $N$ amino acids, 
$\sS \in \{0,1\}^{N\times |V|}$ is a binary matrix, the $i$-th row of $\sS$ is a one-hot vector that indicates the category of the $i$-th amino acid; 
$\bfG = [\bfg_1, \cdots, \bfg_N]^\top \in \RB^{N\times 3}$ represents the coordinates for all the amino acids (coordinates of all $\alpha$ atoms). 
Since rotating or translating $\bfG$ would not change the underlying 3D graph,  
it is challenging to generate such a 3D graph $\bfG$~\cite{satorras2021n}. 
To circumvent the direct generation of 3D amino acids' coordinates $\bfG$, we instead leverage the pairwise distance matrix $\bfD\in \RB_{+}^{N\times N}$, which describes the Euclidean distance between the amino acids. 
That is, the $i,j$-th element of $\bfD$ satisfies $\bfD_{i,j} = ||\bfg_i - \bfg_j||_2$. 
The CDR loop can be represented as $\xx = (\bfD, \sS)$. An illustration of data representation is given in Figure~\ref{fig:data}.

Now, we show that distance matrix $\bfD$ is invariant with respect to translation and rotation on the 3D graph whose coordinates are $\bfG$. That is, rotating and translating $\bfG$ would not change $\bfD$. 
\begin{corollary}
\label{cor:invariance}
Suppose $\bfG'$ are the 3D coordinates of the graph by rotating and translating the original graph whose coordinates are $\bfG$, $\bfD$ and $\bfD'$ are the distance matrix of $\bfG$ and $\bfG'$, respectively, then we have $\bfD = \bfD'$. The proof is given in the Appendix. 
\end{corollary}

\noindent\textbf{Validity constraints}. We define the validity of the generated loop based on empirical domain knowledge about CDR loops \cite{zhang2008tasser,stanfield2014antibody,nowak2016length}. 
Specifically, we define the validity of a generated 3D CDR loop when it satisfies the following two constraints: 
\begin{enumerate}[leftmargin=*]
\item \textbf{Fixed bond length}. The amino acid bonds have relatively fixed length, i.e., the distance between connected amino acids are relatively fixed~\cite{zhang2008tasser}:
\begin{equation}
\label{eqn:bond_length}
\eta_1 \leq \bfD_{i,i+1} \leq \eta_2, \ \text{for}\ i=1,\cdots,N-1. 
\end{equation}
\item \textbf{Open loop}. The shape of CDR is an open loop, where the distance between the first and the last amino acids is within a specific range~\cite{stanfield2014antibody,nowak2016length}, 
\begin{equation}
\label{eqn:open_loop}
\epsilon_1 \leq \bfD_{1,N} \leq \epsilon_2.  
\end{equation}
\end{enumerate}
The setup of $\eta_1,\eta_2, \epsilon_1, \epsilon_2$ is based on domain knowledge~\cite{stanfield2014antibody,nowak2016length} and empirical validation. 
{For example, for CDR H3 loop, we set $\eta_1=3.71\mathring{A}$, $\eta_2=3.88\mathring{A}$, $\epsilon_1=6.5\mathring{A}, \epsilon_2 = 8.5\mathring{A}$}, where Angstrom $\mathring{A}$ is a metric unit of length equal to $10^{-10}$ m. Setups for H1 and H2 loops are given in Section~\ref{sec:implement} in Appendix.

\subsection{Background and Overview}
\label{sec:flow}
\noindent\textbf{Flow model background}. 
The objective of flow model is to learn an invertible model $f_{\theta}$ (with learnable parameter $\theta$) that maps the input data ${\xx} \in \RB^{d} $ into the latent variable ${\bfz} \in \RB^{d}$, i.e., ${\bfz} = f_{\theta}({\xx})$. 
$f_{\theta}$ is invertible, i.e., ${\xx} = f^{-1}_{\theta}({\bfz})$. 
The latent variable has the same size as the input data. 
We suppose the latent variable ${\bfz}$ is drawn from a known prior distribution (e.g., isotropic Gaussian distribution), i.e., ${\bfz}\sim p_{{\bfz}}({\bfz})$. 
% The advantage of flow model is its ability to reconstruct input data exactly due to invertibility, and it directly optimizes the likelihood, instead of optimizing a surrogate, e.g., VAE optimizes the lower bound of likelihood~\cite{kingma2013auto}. 
Using the change-of-variable formula, the log-probability of ${\xx}$ becomes 
\begin{equation}
\label{eqn:loglikelihood}
\log p_{{\xx}}({\xx}) = \log p_{\bfz}({\bfz}) + \log \bigg(\bigg|\det \frac{\partial {\bfz}}{\partial {\xx}} \bigg| \bigg), 
\end{equation}
where $\frac{\partial {\bfz}}{\partial {\xx}}$ is the Jacobian of $f_{\theta}$ at $\xx$, $\det$ is the determinant of a square matrix.  
Recall input data $\xx$ is $\xx = (\bfD, \sS), \bfD \in \RB^{N\times N}, \bfS\in \{0,1\}^{N\times|V|}$. 
\marked{The design of $f$ follows two strategies to learn complex transformations: (i) ``easy logarithm of determinant of the Jacobian'' ($\log \big|\det\frac{\partial \bfz}{\partial \bfx}\big|$) and (ii) ``easy inverse'' ($f^{-1}$)~\cite{dinh2014nice}.}

\noindent\marked{\textbf{Background of affine coupling}. Affine coupling layer is a commonly-used neural architecture in flow model to learn complex transformation, which naturally meets the two criteria above~\cite{dinh2014nice,madhawa2019graphnvp,zang2020moflow}. } 
Specifically, in affine coupling layer, the input $\bfI \in \RB^{P\times Q}$ is decomposed into $\bfI = (\bfI_1, \bfI_2), \bfI_1 \in \RB^{P_1\times Q}, \bfI_2 \in \RB^{(P-P_1)\times Q}$, where $P_1 = \lceil{\frac{P}{2}}\rceil$ (ceiling function) or $P-\lceil{\frac{P}{2}}\rceil$. The mapping is 
\begin{equation}
\label{eqn:affine0}
\bfO_1 = \bfI_1, \ \ \bfO_2 = \bfI_2 \odot \text{sigmoid}(R(\bfI_1)) + T(\bfI_1),     
\end{equation}
where $\odot$ denotes element-wise multiplication. 
Outputs $\bfO_1\in\RB^{P_1\times Q}, \bfO_2\in\RB^{(P-P_1)\times Q}$ has same shape with the inputs $\bfI_1\in\RB^{P_1\times Q}, \bfI_2\in\RB^{(P-P_1)\times Q}$, respectively. 
$R(\cdot), T(\cdot): \RB^{P_1\times Q}\xrightarrow[]{} \RB^{(P-P_1)\times Q}$ share the same neural architectures but different parameters. 
The sigmoid function is applied element-wisely. 
The logarithm of the Jacobian determinant is easy to evaluate, 
\begin{equation}
\label{eqn:jacob_affine}
\begin{aligned}
&\log \bigg| \det \frac{\partial \bfO}{\partial \bfI} \bigg|
% \\ &
= \log \bigg| \det 
\begin{pmatrix}
\mathbb{I} & 0 \\ 
0 & \text{sigmoid}(R(\bfI_1)) \\
\end{pmatrix} 
\bigg| = \sum\nolimits_{j=1}^{(P-P_1)Q} \log \text{sigmoid}(R(\bfI_1))_j. \\ 
\end{aligned}
\end{equation}
where $\mathbb{I}$ is the identity matrix. 
The inverse mapping is 
\begin{equation}
\label{eqn:inverse_affine}
\begin{aligned}
\bfI_1 = \bfO_1, \ \ \bfI_2 = \big(\bfO_2 - T(\bfI_1) \big) / \text{sigmoid}\big(R(\bfI_1)\big). 
\end{aligned}
\end{equation}
Multiple affine coupling layers are usually concatenated (e.g., $\calF_1$, $\cdots$, $\calF_K$ in Eq.\ref{eqn:distanceflow}). 
Consecutive affine coupling layers swap the elements in $\bfI_1$ and $\bfI_2$ so that every element occurs in $\bfI_1$ and $\bfI_2$ in turn.

\noindent\textbf{Overview of \mname}. 
\mname generates 3D CDR loop $\xx$ in two phases. 
The first phase generates the distance matrix $\bfD$.
In the second phase, based on $\bfD$, we predict the amino acids for all the nodes, denoted $\bfS$. 
Specifically, the latent variable ${\bfz}$ is decomposed as ${\bfz} = ({\bfz}_{\bfD}, {\bfz}_{\sS|\bfD})$, the invertible model $f=(f_{\bfD}, f_{\sS|\bfD})$\footnote{$f$ should be formally written as $f_\theta$. We omit $\theta$ for simplicity.} has two components. 
The two latent variables are ${\bfz}_{\bfD} = f_{\bfD}(\bfD), \bfz_{\bfD} \in \RB^{N\times N}$, and ${\bfz}_{\sS|\bfD} = f_{\sS|\bfD}(\sS|\bfD), \bfz_{\sS|\bfD} \in \RB^{N\times|V|}$. 
The Jacobian of ${\bfz} = f({\xx})$ is decomposed as 
% \begin{equation}
\[
\frac{\partial {\bfz}}{\partial {\xx}} = \frac{\partial ({\bfz}_{\bfD}, {\bfz}_{\sS|\bfD})}{\partial (\bfD, \sS)} = 
\begin{bmatrix}
\frac{\partial {\bfz}_{\bfD}}{\partial \bfD} & \frac{\partial {\bfz}_{\bfD}}{\partial \sS} \\
\frac{\partial {\bfz}_{\sS|\bfD}}{\partial \bfD} & \frac{\partial {\bfz}_{\sS|\bfD}}{\partial \sS} \\ 
\end{bmatrix}
= \begin{bmatrix}
\frac{\partial {\bfz}_{\bfD}}{\partial \bfD} & 0 \\
\frac{\partial {\bfz}_{\sS|\bfD}}{\partial \bfD} & \frac{\partial {\bfz}_{\sS|\bfD}}{\partial \sS} \\ 
\end{bmatrix}
\]
% \end{equation}
The logarithm Jacobian determinant of $f_{\theta}$ can be decomposed as $\log \big|\det \frac{\partial \bfz}{\partial \bfx}\big| = \log \big| \det \frac{\partial ({\bfz}_{\bfD}, {\bfz}_{\sS|\bfD})}{\partial (\bfD, \sS)} \big| = \log \big|\det \frac{\partial {\bfz}_{\bfD}}{\partial \bfD} \big| + \log \big|\det \frac{\partial {\bfz}_{\sS|\bfD}}{\partial \sS} \big|$. 
The flow model is split into two parts: (1) distance matrix flow (Section~\ref{sec:distance}); (2) conditional amino acid flow (conditioned on distance matrix) (Section~\ref{sec:amino}). 
Section~\ref{sec:constraint} discusses how flow model learns constraints; 
Section~\ref{sec:coordinate_generation} describes how to generate 3D coordinates based on both distance matrix and constraints. 
Section~\ref{sec:learning} (Appendix) describes the learning and inference process. 
The pipeline is illustrated in Fig.~\ref{fig:pipeline}. 
We slightly abuse the mathematical notations for simplicity.

\subsection{Distance Matrix Flow}
\label{sec:distance}

This section elaborates on distance matrix flow $f_{\bfD}(\bfD)$. We illustrate the neural architecture in Figure~\ref{fig:pipeline}. 
% following the order: $G \xrightarrow{(1)} D \xrightarrow{(2)} \bfz_{G} \xrightarrow{(3)} D \xrightarrow{(4)\text{}} \tilde{G}$. 
% \noindent\textbf{(1): $G\xrightarrow[]{} D$.} Distance matrix is easy to compute given the graph $\bfG$. 
% \noindent\textbf{(2): $D\xrightarrow[]{} \bfz_G$}. 
To build invertible mappings, we concatenate $K$ affine coupling layers (Eq.~\ref{eqn:affine0}, denoted $\calF_1, \cdots, \calF_K$), which is a popular technique in flow model~\cite{madhawa2019graphnvp,zang2020moflow}, 
\begin{equation}
\label{eqn:distanceflow}
f_{\bfD} = \calF_1 \circ \calF_2 \circ \cdots \circ \calF_K. 
\end{equation}
The logarithm of Jacobian determinant of $f_{\bfD}$ is 
\begin{equation}
\label{eqn:fd_jac}
\log \bigg| \det \frac{\partial f_{\bfD}}{\partial \bfD} \bigg|
= \sum\nolimits_{k=1}^{K} \log \bigg| \det \frac{\partial \calF_k}{\partial Y_k} \bigg|. 
\end{equation}
where $Y_k$ is the input of $k$-th affine coupling layer, and also the output of ($k$-1)-th affine coupling layer, $\big| \det \frac{\partial \calF_k}{\partial Y_k} \big|$ is given in Eq.~\eqref{eqn:jacob_affine}. 
The inverse flow $f_{\bfD}^{-1}$ reverses the order to apply the $K$ affine coupling layers (Eq.~\ref{eqn:distanceflow}): 
\begin{equation}
\label{eqn:inverse_distance}
f^{-1}_{\bfD} = \calF_K^{-1} \circ \calF_{K-1}^{-1} \circ \cdots \circ \calF_1^{-1}. 
\end{equation} 
The inverse of affine coupling ($\calF^{-1}_k$) is given in Eq.~\eqref{eqn:inverse_affine}. 
The mapping of an affine coupling layer is given in Eq.~\ref{eqn:affine0}. 
As mentioned, scale/transformation function $R(\cdot)/T(\cdot)$ (Eq.~\ref{eqn:affine0}) have the same neural architecture: to capture the local relationship between the distance matrix, we leverage a two-dimensional convolutional neural network (CNN2D) followed by a batch normalization (BN) layer and ReLU activation, shown in Fig~\ref{fig:pipeline}.

% \begin{figure}[htb]
% \centering
% \includegraphics[width=\columnwidth]{figure/architecture.png}
% \caption{The neural architecture of $f = (f_{\bfD}, f_{\sS|\bfD})$. 
% As described in Eq.~\eqref{eqn:affine}, $\bfI_1, \bfI_2$ are inputs of affine coupling layer while $\bfO_1,\bfO_2$ are outputs, $R()/T()$ are scale and transformation function, respectively.  
% }
% \label{fig:architecture}
% \end{figure}

\subsection{Conditional Amino Acid Flow}
\label{sec:amino}

This section describes the conditional amino acid flow $f_{\sS|\bfD}(\sS|\bfD)$, which shares similarity with the distance matrix flow in Section~\ref{sec:distance}. The main difference is the neural architecture of $R(\cdot)/T(\cdot)$, as shown in Figure~\ref{fig:pipeline}. 
% We provide the details as follows. 

Conditional amino acid flow model $f_{\sS|\bfD}$ concatenates $L$ affine coupling layers, denoted $\calH_1, \cdots, \calH_L$
\begin{equation}
\label{eqn:amino_flow}
f_{\sS|\bfD} = \calH_1 \circ \calH_2 \circ \cdots \circ \calH_L. 
\end{equation}
Similar to Eq.~\eqref{eqn:fd_jac} and~\eqref{eqn:inverse_distance}, the logarithm of Jacobian determinant of $f^{}_{\sS|\bfD}$ and the inverse are 
\begin{equation}
\label{eqn:inverse_amino}
\begin{aligned}
& \log \bigg| \det \frac{\partial f_{\sS|\bfD}}{\partial \sS} \bigg| = \sum\nolimits_{l=1}^{L} \log \bigg| \det \frac{\partial \calH_l}{\partial Y_l} \bigg|, \ \ \ \  
% \\ & 
f^{-1}_{\bfS|\bfD} = \calH_{L}^{-1} \circ \calH_{L-1}^{-1} \circ \cdots \circ \calH_{1}^{-1}. 
\end{aligned}
\end{equation}
where $Y_l$ is the input of $l$-th affine coupling layer and also the output of ($l$-1)-th layer. 
For $\calH_{l}$ ($l$=1,$\cdots$,$L$), the mapping of an affine coupling layer is 
$\bfO_1 = \bfI_1, \bfO_2 = \bfI_2 \odot \text{sigmoid}\big(R(\bfI_1 |\bfD) \big) + T(\bfI_1 |\bfD)$ (Eq.~\ref{eqn:affine0}). 
$\big| \det \frac{\partial \calH_l}{\partial Y_l} \big|$ and $\calH_{1}^{-1}$ are in Eq.~\eqref{eqn:jacob_affine} and~\eqref{eqn:inverse_affine}, respectively. 
Consecutive affine coupling layers ($\calH_1, \cdots, \calH_K$) swap the elements in $\bfI_1$ and $\bfI_2$. 

Scale/transformation functions $R(\cdot)/T(\cdot)$ (Eq.~\ref{eqn:affine0}) have same architecture. 
The architecture consists of (1) wGNN (weighted-distance graph neural network, Eq.\ref{eqn:wgnn}); (2) Batch Normalization (BN), and (3) ReLU activation. 
The traditional graph neural networks (GNNs), e.g., graph convolutional network~\cite{kipf2016semi}, graph attention network~\cite{velivckovic2017graph}, pass the message between nodes through binary adjacency matrix.
In contrast, in our scenario, the CDR loop is a single chain and is folded into the 3D structure. 
To characterize the interaction between nodes, we design weighted-distance graph neural network to incorporate weighted edge, denoted wGNN, 
\begin{equation}
\label{eqn:wgnn}
\text{wGNN}(\bfI_1 | \bfD) =  \tilde{\bfD} (\mathbf{M}(\bfI_1) \odot \bfI) \mathbf{W}, 
\end{equation}
where $\mathbf{W} \in \RB^{|V| \times |V|}$ are the learnable weight matrix;
$\mathbf{M}(\bfI_1) \in \{0,1\}^{N\times |V|}$ is a binary mask to select a partition $\bfI_1$ from $\bfI\in \RB^{N\times |V|}$; $\tilde{\bfD} \in \RB^{N \times N} $ is weighted edge matrix; 
the $(i,j)$-th element $\tilde{\bfD}_{i,j} = \exp(-\xi \bfD_{i,j})$ measures the interaction strength between $i$-th and $j$-th nodes, $\xi>0$ is the hyperparameter. Intuitively, when the $i$-th and $j$-th nodes are closer, the interaction strength is stronger. In sum, $\tilde{\bfD}$ replaces the binary adjacency matrix in the vanilla graph neural network. 
% % \tf{actnorm2d, 4.2.3}
% To guarantee the numerical stability, following~\cite{kingma2018glow}, we leverage a variant of invertible two-dimensional activation normalization layer for the amino acid matrix, denoted as \textit{\underline{actnorm2d}}. % (activation normalization for 2D matrix), 
% To normalize each row, i.e., the feature dimension for each node, over a batch of 2-dimensional
% amino acid matrices. 
% Given the mean $\mu\in\RB^{n\times 1}$ and the standard deviation $\sigma^2\in \RB^{n\times 1}$, for each row dimension, the normalized input follows $\hat{A} = \frac{A-\mu}{\sqrt{\sigma^2 + \epsilon}}$,
% where $\epsilon > 0$ is small constant to guarantee the invertible transformation. 
% The reverse transformation is 
% $A = \hat{A} * \sqrt{\sigma^2 + \epsilon} +\mu$, and the logarithmic Jacobian determiant is:
% \begin{equation}
% \log \big|\det\frac{\partial \text{actnorm2d}}{\partial X} \big| = \frac{k}{2} \sum_{i=1}^{n} |\log(\sigma_i^2 + \epsilon)|
% \end{equation}

\subsection{Constraint Learning}
\label{sec:constraint}
Constraint learning encourages the flow model to learn the validity constraints of the CDR loop. 
To proceed, we design a \textit{constraint loss} function, a function of the generated distance matrix $\bfD$. It  penalizes the distance matrix that does not satisfy the validity constraints. 
However, the constraints in Eq.~\eqref{eqn:bond_length} and~\eqref{eqn:open_loop} are hard constraints and not differentiable w.r.t. the distance matrix. 
To bridge the gap, we design the following \textit{constraint surrogate function} $H(y; a, b, \delta)$ that transforms the hard constraints into soft constraints. 

The \textit{constraint surrogate function} $H(y; a, b, \delta)$ is a function of $y$ (a scalar), where $a, b, \delta$ are all scalars and satisfy $a < b$ and $\delta > 0$. $H$ is designed to encourage the variable $y$ to be between scalar $a$ and $b$. 
To make $H$ differentiable everywhere and enable the continuous optimization process, 
we borrow the idea from Huber loss~\cite{yi2017semismooth}. 
The basic idea of Huber loss comes from the fact that square loss is sensitive to outliers but differentiable everywhere, whereas L1 loss is more robust to outliers but not differentiable at 0. 
Combining the advantage of both, the Huber loss function is quadratic for small values (square loss), and linear for large values (L1 loss). 
In our scenario, as long as the scalar $y$ is between $a$ and $b$, the constraints are satisfied with  the loss equal to 0. 
Thus, we design $H(y; a,b,\delta)$ as 
\begin{equation}
\label{eqn:huber}
\begin{aligned}
H(y; a, b, \delta) = 
\left\{
\begin{aligned}
& \delta(-y+a-\frac{1}{2}\delta), & & y \leq a - \delta \\ 
& \frac{1}{2}(y-a)^2, & & a - \delta < y \leq a, \\
&0, & & a < y\leq b, \\
& \frac{1}{2}(y-b)^2, & & b < y \leq b + \delta, \\  
& \delta(y-b-\frac{1}{2}\delta), & & b+\delta < y. \\
\end{aligned}
\right.
\end{aligned}
\end{equation}

It has equal values and slopes of the different sections at the four connection points ($a-\delta, a, b, b+\delta$), so it is continuous and differentiable everywhere. 
% Figure~\ref{fig:H} illustrates $H$ function. 
When $a=b$, $H$ loss function reduces to Huber loss function. 
Thus, the hard constraints in Eq.~\eqref{eqn:bond_length} and~\eqref{eqn:open_loop} can be transformed into surrogate functions as follows 
\begin{equation}
\begin{aligned}
h_1(\bfD) = \sum\nolimits_{i=1}^{N-1} H(\bfD_{i,i+1}; \eta_1, \eta_2, \eta_3), \ \ \ \ \ h_2(\bfD) = H(\bfD_{1,N}; \epsilon_1, \epsilon_2, \epsilon_3). \\ 
\end{aligned}
\end{equation}
where $\eta_1/\eta_2$ and $\epsilon_1/\epsilon_2$ are defined in Eq.~\eqref{eqn:bond_length} and~\eqref{eqn:open_loop}, respectively. We also have $\eta_3 = \eta_2 - \eta_1$ and $\epsilon_3=\epsilon_2-\epsilon_1$. 
\begin{algorithm}[h!]
\caption{\mname} 
\label{alg:main}
\begin{algorithmic}[1]
\STATE \textbf{Input}: data ${\xx}=\{\sS,\bfD\}$; Training iteration $T$. 
\STATE \textbf{Output}: $\widetilde{{X}^{(1)}}, \widetilde{{X}^{(2)}}, \cdots$. 
\STATE \#\#\#\# Learning \#\#\#\# 
\FOR{$t=1,2,\cdots,T$} 
\STATE Maximize log-likelihood $\log p_{\xx}(\xx)$(Eq.\ref{eqn:loglikelihood})(Sec\ref{sec:distance}-\ref{sec:amino})
\STATE Minimize constraint loss $\widetilde{\calL_{\text{con}}}$  (Eq.~\ref{eqn:constraint_obj}) (Sec~\ref{sec:constraint}). 
\ENDFOR 
\STATE \#\#\#\# Inference \#\#\#\#
\FOR{$i=1,2,\cdots,$}
\STATE Draw latent variable ${\bfz}\sim p_{\bfz}({\bfz})$, and ${\bfz} = ({\bfz}_{\bfD}, {\bfz}_{\sS|\bfD})$. 
\STATE Generate distance matrix: $\bfD = f_{\bfD}^{-1}({\bfz}_{\bfD})$ (Eq.~\ref{eqn:inverse_distance}). 
\STATE Generate amino acid matrix: $\sS = f_{\sS|\bfD}^{-1}({\bfz}_{\sS|\bfD}; \bfD)$ (Eq.~\ref{eqn:inverse_amino}). 
\STATE Constrained 3D coordinates generation: $\bfG$ (Eq.~\ref{eqn:graph_generation}) (Sec~\ref{sec:coordinate_generation}). 
\STATE $\widetilde{{X}^{(i)}} = (\bfS, \bfG)$. 
\ENDFOR 
\end{algorithmic}
\end{algorithm}
$h_1(\cdot)$ and $h_2(\cdot)$ consider $\bfD_{1,N}$ and $\bfD_{i,i+1}$ ($i=1,\cdots,N-1$), respectively. 
On the other hand, based on the validity constraint (Eq.\ref{eqn:bond_length}) and triangle inequality, we induce some constraints on $\{\bfD_{i,j}\}$ ($i,j\in\{1,\cdots,N\}$) as follows, 
\begin{equation}
\label{eqn:smooth}
\begin{aligned}
& |\bfD_{i,j} - \bfD_{(i+1),j}| \leq \bfD_{i,(i+1)} \leq \eta_2, \ \ \ |\bfD_{i,j} - \bfD_{i,(j+1)} | \leq \bfD_{j,(j+1)} \leq \eta_2, 
\\ &
\text{for}\ i,j\in\{1,2,\cdots, N-1\},  
\end{aligned}
\end{equation}
where $\eta_2$ is defined in Eq.~\eqref{eqn:bond_length}. 
It indicates that the elements of $\bfD$ are smooth, we design the following \textit{constraint loss} to encourage the smoothness,  
\begin{equation}
\begin{aligned}
& h_3(\bfD)
=  \sum_{i=1}^{N-1}\sum_{j=1}^{N-1} ( \bfD_{i,j} - \bfD_{(i+1),j} )^2  + ( \bfD_{i,j} - \bfD_{i,(j+1)} )^2.  
\end{aligned}
\end{equation}

We design \textit{constraint loss} function as follows, which is a function of distance matrix $\bfD$, 
\begin{equation}
\label{eqn:constraint_loss}
\begin{aligned}
& \ h(\bfD) = \xi_1 h_1(\bfD) + \xi_2 h_2(\bfD) + \xi_3 h_3(\bfD), \\ 
\end{aligned}
\end{equation}
where $\xi_1, \xi_2, \xi_3 > 0$ are the hyperparameters that control the weights of various terms. 
When generating new distance matrix, we draw latent variable ${\bfz}_{\bfD}$ from isotropic Gaussian distribution $p_{{\bfz}_{\bfD}}(\cdot)$, the learning objective is designed as expectation of $h(\bfD)$ under the distribution of $\bfz_{\bfD}$ 
\begin{equation}
\begin{aligned}
\calL_{\text{con}} (\theta) = \EB_{\big[{\bfz}_{\bfD} \sim p_{{\bfz}_{\bfD}}(\cdot), \bfD=(f_{\theta})_{\bfD}^{-1}({\bfz}_{\bfD}) \big]} h(\bfD).  
\end{aligned}
\end{equation}
We use Monte Carlo sampling to get a noisy but unbiased estimate of the learning objective, 
\begin{equation}
\label{eqn:constraint_obj}
\begin{aligned}
& \widetilde{\calL_{\text{con}} (\theta)} = \frac{1}{M} \sum\nolimits_{i=1}^{M} h(\bfD^{(i)}), \  \text{where}\  {\bfz}^{(i)}_{\bfD} \overset{\text{i.i.d.}}{\sim} p_{{\bfz}_{\bfD}}(\cdot), \\ & i=1,\cdots,M, \ \ \ \ \ \ \bfD^{(i)} = (f_{\theta})_{\bfD}^{-1}({\bfz}^{(i)}_{\bfD}), 
\end{aligned}
\end{equation}
where $M$ is the number of Monte Carlo samples. 
Gradient descent is leveraged to minimize the noisy objective $\widetilde{\calL_{\text{con}} (\theta)}$. 
We alternatively optimize log-likelihood in Eq.~\eqref{eqn:loglikelihood} and constraint loss in Eq.~\eqref{eqn:constraint_loss}. 
Constraint learning is only conducted during learning and not utilized in inference.

\subsection{Constrained 3D Coordinates Generation}
\label{sec:coordinate_generation}

This section discusses how to transform the generated distance matrix $\bfD$ into 3D coordinates $\bfG$ that satisfy validity constraints simultaneously. 
Specifically, given the generated distance matrix $\bfD \in \RB^{N\times N}$, we build the 3D graph via optimizing the coordinates of all the amino acids, i.e., $\bfG = [\bfg_1, \bfg_2, \cdots, \bfg_N]^{\top} \in \RB^{N\times 3}$. 
We optimize the amino acids' coordinates by minimizing the following objective under the validity constraints given in Eq.~\eqref{eqn:bond_length} and~\eqref{eqn:open_loop}: 
\begin{equation}
\label{eqn:graph_generation}
\begin{aligned}
& \underset{\bfg_1,\cdots,\bfg_N\in\RB^3}{\arg\min} \sum\nolimits_{i=1}^{N} \sum\nolimits_{j=1,j\neq i}^{N} \big| \Vert \bfg_i - \bfg_j \Vert_2^2 - \bfD_{i,j}^2 \big|, \\ 
& \text{s.t.} \  \eta_1 \leq ||\bfg_i - \bfg_{i+1} ||_2 \leq \eta_2, \  \epsilon_1 \leq || \bfg_1 - \bfg_{N} ||_2 \leq \epsilon_2, \\ 
\end{aligned}
\end{equation}
where $\eta_1, \eta_2$ are defined in Eq.~\eqref{eqn:bond_length}, $\epsilon_1, \epsilon_2$ are defined in Eq.~\eqref{eqn:open_loop}, $||\cdot||_2$ represents ${l}_2$ norm of a vector.

Then, in order to simplify the optimization process, we used the \textit{constraint surrogate function} $H(\cdot)$ (Eq.~\eqref{eqn:huber}) to turn the hard constraints in Eq.~\eqref{eqn:graph_generation} into soft constraints. 
Then we augment the learning objective and reformulate an unconstrained optimization problem as follows,  
\begin{equation}
\label{eqn:unconstrained_optimization}
\begin{aligned}
 & \underset{\bfg_1,\cdots,\bfg_N\in\RB^3}{\arg\min} \sum\nolimits_{i=1}^{N} \sum\nolimits_{j=1,j\neq i}^{N} \big| \Vert \bfg_i - \bfg_j \Vert_2^2 - \bfD_{i,j}^2 \big| \\
& + \lambda_1 \sum_{i=1}^{N-1} H(||\bfg_{i} - \bfg_{i+1}||_2; \eta_1, \eta_2, \eta_3)
% \\ & 
+ \lambda_2 H(||\bfg_1 - \bfg_N||_2; \epsilon_1, \epsilon_2, \epsilon_3),  
\end{aligned}
\end{equation}
where $H(\cdot)$ is the loss function defined in Eq.~\eqref{eqn:huber}; hyperparameters $\lambda_1, \lambda_2 \gg 1$ emphasize the importance of the constraints to make sure that they are satisfied, hyperparameters $\eta_3$ and $\epsilon_3$ are set to $\eta_3 = \eta_2 - \eta_1$, $\epsilon_3 = \epsilon_2 - \epsilon_1$.

To optimize Eq.~\eqref{eqn:unconstrained_optimization}, we leverage random coordinate descent~\cite{necoara2014random} to update $\bfg_1, \cdots, \bfg_N$ (i.e., randomly selecting a $\bfg_i$ from $\{\bfg_1, \cdots, \bfg_N\}$ to update and fixing the remaining ones). 
$H(y; a,b,\delta)$ is convex w.r.t. $y$. 
Thus, when we update each $\bfg_i$ ($i=1,\cdots,N$) in coordinate descent, when fixing other $\bfg$ variables, Eq.~\eqref{eqn:unconstrained_optimization} is convex w.r.t. $\bfg_i$, which makes the optimization efficient. 
Worth mentioning that the constrained 3D coordinates generation is not conducted during the learning procedure and only during inference. 
Algorithm~\ref{alg:main} summarizes the essential steps.

\section{Experiments}
\label{sec:experiment}

\subsection{Experimental Setup}
\label{sec:setup}

\noindent\underline{\textbf{Dataset}}. (1) \textbf{SAbDab} (structural antibody database) (\url{http://opig.stats.ox.ac.uk/webapps/newsabdab/sabdab/}) is a database containing 5,494 antibodies structures~\cite{dunbar2014sabdab}. (2) \textbf{CoVAbDab} (coronavirus antibody database)~\cite{raybould2021cov} (\url{http://opig.stats.ox.ac.uk/webapps/covabdab/}) documents 2,942 published binding antibodies to coronaviruses, the 3D structures are not available. Data split strategies are given in Section~\ref{sec:data}.

\noindent\underline{\textbf{Metrics}}. 
(1) \textbf{Protein Perplexity (PPL)} measures protein likeness on amino acid sequence~\cite{ingraham2019generative};
(2) \textbf{Root-Mean-Square Deviation (RMSD)} measures the alignment between generated 3D conformations and reference 3D conformation; 
(3) \textbf{Validity Rate (VR)} is the percentage of valid (satisfying Eq.\ref{eqn:bond_length},~\ref{eqn:open_loop}) CDR loops in all the generated CDRs; 
\noindent (4) \textbf{Diversity (Div)} measures the diversity of generated CDR loops on the amino acid sequence level. 
\noindent (5) \textbf{SARS-CoV-2 Neutralization Ability}(used in \cite{jin2021iterative}) is a well-trained machine learning model to evaluate the neutralization ability of a CDR-H3 sequence to either SARS-CoV-1/2 viruses. Detailed definitions are in Appendix.

\noindent\underline{\textbf{Baselines}}.
\textbf{(1) Reference} evaluates the results of CDR loops in the test set; 
\textbf{(2) LSTM (long short-term memory)}~\cite{saka2021antibody,akbar2021silico}; 
\textbf{(3) GA (genetic algorithm)}~\cite{liu2017computational}; 
\textbf{(4) IR-GNN} (Iterative refinement graph neural network)~\cite{jin2021iterative}; \textbf{(5) AR-GNN} (auto-regressive graph neural network)~\cite{you2018graphrnn,luo20213d}. 
\textbf{(6) \mname \ - constraint} is the variant of \mname that does not consider validity constraint. 
LSTM, IR-GNN and AR-GNN are deep generative models while GA is an evolutionary algorithm. 
LSTM is  unable to model 3D structures hence the results on RMSD and VR are unavailable. 
All the approaches exhibit state-of-the-art performance on antibody design~\cite{saka2021antibody,akbar2021silico,jin2021iterative}. 
More details are given in Appendix.

\begin{table}[tb]
\small 
\centering
\caption{CDR loop (H1, H2, H3) generation results on SAbDAb. }
\resizebox{0.62\columnwidth}{!}{
\begin{tabular}{c|c|c|c|c|c}
\toprule[1pt]
 & Method & PPL ($\downarrow$) & RMSD ($\downarrow$) & VR ($\uparrow$) & Div ($\uparrow$) \\ 
\midrule[1pt]
\multirow{7}{*}{H1} & Reference & 8.04$\pm$0.77 & 0.0 & 100.0\% & 0.501 \\ 
 & LSTM & 10.01$\pm$1.21 & - & - & 0.550 \\ 
 & GA & 10.45$\pm$1.42 & 1.93$\pm$0.76 & 44.0\% & 0.631 \\ 
 & AR-GNN & 9.53$\pm$1.58 & 1.93$\pm$0.72 & 61.8\% & 0.625 \\ 
 & IR-GNN & 9.10$\pm$1.37 & 1.69$\pm$0.47 & 88.6\% & \textbf{0.681} \\  
 & \mname \ - constraint & 9.32$\pm$1.21 & 1.92$\pm$0.75 & 71.4\% & 0.635 \\ 
 & \mname & \textbf{8.68$\pm$0.92} & \textbf{1.28$\pm$0.45} & \textbf{96.3\%} & 0.645 \\ 
\midrule[1pt]
\multirow{7}{*}{H2} & Reference & 8.50$\pm$1.10 & 0.0 & 100.0\% & 0.697 \\ 
& LSTM & 10.90$\pm$1.65 & - & - & 0.634 \\ 
& GA & 10.20$\pm$1.20 & 1.95$\pm$0.49 & 33.3\% & 0.543 \\ 
& AR-GNN & 10.54$\pm$1.63 & 1.69$\pm$0.30 & 34.5\% & 0.667 \\ 
& IR-GNN & 9.63$\pm$1.40 & 1.12$\pm$0.17 & 86.9\% & 0.602 \\  
& \mname \ - constraint & 9.96$\pm$1.89 & 1.70$\pm$0.30 & 49.5\% & 0.672 \\ 
& \mname & \textbf{9.23$\pm$1.32} & \textbf{0.99$\pm$0.11} & \textbf{92.8\%} & \textbf{0.674} \\ 
\midrule[1pt]
\multirow{7}{*}{H3} 
& Reference & 9.82$\pm$0.48 & 0.0 & 100.0\% & 0.743 \\  
& LSTM & 12.35$\pm$0.93 & - & - & 0.731 \\ 
& GA & 12.75$\pm$1.29 & 4.33$\pm$1.70 & 13.4\% & 0.713\\ 
& AR-GNN & 13.01$\pm$1.13 & 3.80$\pm$1.69 & 25.5\% & 0.748 \\
& IR-GNN & 11.43$\pm$0.65 & 3.08$\pm$0.74 & 78.4\% & 0.753 \\ 
& \mname \ - constraint & 10.96$\pm$0.80 & 3.21$\pm$0.98 & 59.1\% & 0.769 \\ 
& \mname & \textbf{10.81$\pm$0.47} & \textbf{2.50$\pm$0.50} & \textbf{91.0\%} & \textbf{0.780} \\ 
\bottomrule[1pt]
\end{tabular}}
\label{table:denovo}
\end{table}

\begin{figure}
\centering
\includegraphics[width=\columnwidth]{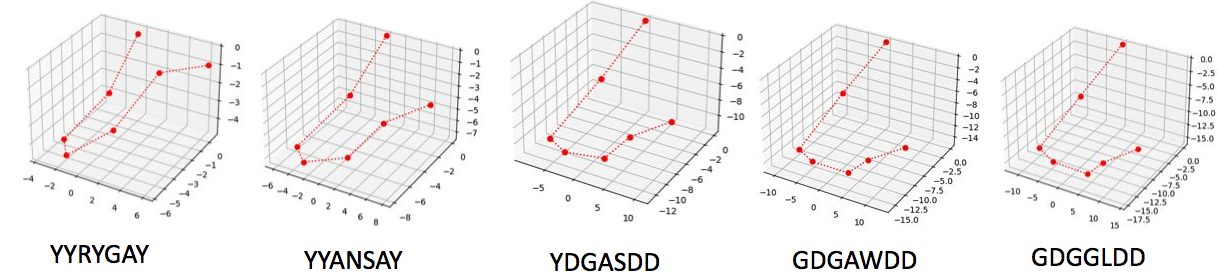}
\caption{Visualization of interpolation between two CDR H3 loops. For both the 3D geometric structure and amino acid sequence, the changing trajectories are smooth. 
}
\label{fig:visualize}
\end{figure}

\begin{table}[tb]
\small 
\centering
\caption{SARS-CoV-2 neutralization optimization results (H3 sequence) on CoVAbDab. }
\resizebox{0.5\columnwidth}{!}{
\begin{tabular}{c|c|c|c}
\toprule[1pt]
 Method & PPL ($\downarrow$) & Neutralization ($\uparrow$) & Div ($\uparrow$) \\ 
 \midrule 
 Reference & 9.23$\pm$0.49 & 0.942$\pm$0.026 & 0.838 \\ 
 \midrule 
 LSTM & 10.64$\pm$0.93 & 0.710$\pm$0.102 & 0.747 \\ 
 GA & 11.34$\pm$1.19 & 0.812$\pm$0.075 & 0.783 \\ 
 AR-GNN & 10.12$\pm$0.81 & 0.801$\pm$0.025 & 0.680 \\ 
 IR-GNN & 9.98$\pm$0.52 & 0.832$\pm$0.024 & 0.734 \\ 
 \mname & \textbf{9.74$\pm$0.47} & \textbf{0.854$\pm$0.011} & \textbf{0.793} \\ 
\bottomrule[1pt]
\end{tabular}}
\label{table:optimization}
\end{table}

\subsection{Results and Analysis}
\label{sec:results}

\noindent\textbf{\textit{De novo} design}. We report the performance of \textit{de novo} CDR H1, H2, and H3 generations in Table~\ref{table:denovo}. For PPL and RMSD, we report the average value and standard deviations of all the data points in the test set. 
We find that (i) in all three tasks, our method achieves over 90\% validity rate (96.3\% on H1, 92.8\% on H2, 91.0\% on H3) and obtains the best performance among all the methods in both sequences (PPL, perplexity) and geometric graph metrics (RMSD \& validity rate). 
Compared to IR-GNN (the best baseline), the relative improvements are up to 5.4\% in PPL (perplexity) (10.81 v.s. 11.43, H3), 24.3\% in RMSD (1.28 v.s. 1.69, H1), and 16.0\% in VR (validity rate) (91.0\% v.s. 78.4\%, H3); 
(ii) among all the three tasks (H1, H2, H3), almost all the methods get the highest perplexity, RMSD, and lower validity in the H3 generation task, thus validating the fact that CDR H3 loops have the highest variability and most challenging to design~\cite{regep2017h3,jin2021iterative}; 
(iii) removing constraint learning from \mname (``\mname \ - constraint'' v.s. \mname) significantly degrades the performance, especially in geometric graph metrics (RMSD and validity rate (VR)). For example, in H3 loop generation (Table~\ref{table:denovo}), it leads to 1.4\%, 22.1\%, and 35.1\% relative drops in PPL, RMSD, and VR, respectively. 
It validates that learning constraint plays a crucial role in geometric graph generation, especially generating a valid CDR loop; (iv) \mname achieves the highest diversity in H2 \& H3 design tasks, and second highest diversity in H1 generation, indicating our method is able to generate diverse amino acid sequences.

\noindent\textbf{Visualization of continuous latent variable}. 
During the generation process, the continuous latent variable $z$ controls the data space $x$ in the flow model, i.e., $x = f^{-1}(z)$. 
Following~\cite{jin2018junction,madhawa2019graphnvp,zang2020moflow}, we visualize a linear interpolation between two CDR loops to show their changing trajectory. 
Specifically, given two CDR loops $x_1$ and $x_2$, we get the latent variables $z_1=f(x_1)$ and $z_2=f(x_2)$. 
Then we get multiple $z$s using linear interpolation between $z_1$ and $z_2$. 
We visualize these $\tilde{z}$ via $\tilde{x}=f^{-1}(\tilde{z})$. 
We show the changing trajectory of the generated CDR loops in Fig~\ref{fig:visualize} and observe a smooth trajectory of both 3D coordinates and amino acid sequence when varying the latent variable.

\noindent\textbf{Optimization}. The SARS-CoV-2 neutralization ability optimization results on CoVAbDab data are reported in Table~\ref{table:optimization}. 
3D graphs are not available in CoVAbDab data, hence we are unable to model the 3D graph. 
For PPL (perplexity) and neutralization (SARS-CoV-2 neutralization ability), we report the average and standard deviations. 
We find that \mname obtains the lowest perplexity (9.74 on average), highest neutralization ability (0.854 on average), and highest diversity (0.793) among all the compared machine learning methods.

\section{Conclusion}
\label{sec:conclusion}

This paper proposed \mname model to generate 3D antibody CDR loops. 
Specifically, our method first builds the distance matrix of amino acids, and then constructs the amino acid sequence conditioned on the distance matrix. 
Also, to enhance the geometric validity of the generated CDR loops, we design constraint learning and constrained 3D coordinate generation to impose the 3D geometric constraints. 
Empirical studies validate \mname's superiority.  
\section*{Acknowledgment}
This work is supported by the Fundamental Research Program of Shanxi Province (No.20210302123025) and the Joint Funds of the Natural Science Foundation of China (No.U21A20524).

\bibliographystyle{unsrt}
\bibliography{ref}

\clearpage  
\appendix

\tableofcontents

\section*{Appendix}

\section{Overview of Appendix}

The supplementary materials are organized as follows. 
Table~\ref{table:notation} lists all the important mathematical notations. 
Section~\ref{sec:related} reviews the existing literature. 
Section~\ref{sec:proof} proves Corollary~\ref{cor:invariance}. 
We describe the learning and inference process in Section~\ref{sec:learning}. 
Then we elaborate on the experimental setup, including data and preprocessing in Section~\ref{sec:data}, evaluation metrics in Section~\ref{sec:metric}, and baseline methods in Section~\ref{sec:baseline}. 
Implementation details are given in Section~\ref{sec:implement}. 
Section~\ref{sec:sensitive} conducts sensitivity analysis on some key hyperparameters. 
Section~\ref{sec:casestudy} shows a case study. 
We also upload the code repository in the supplementary materials\footnote{Code repository is available in the supplementary materials. }.

\begin{figure}
\centering
\includegraphics[width=\columnwidth]{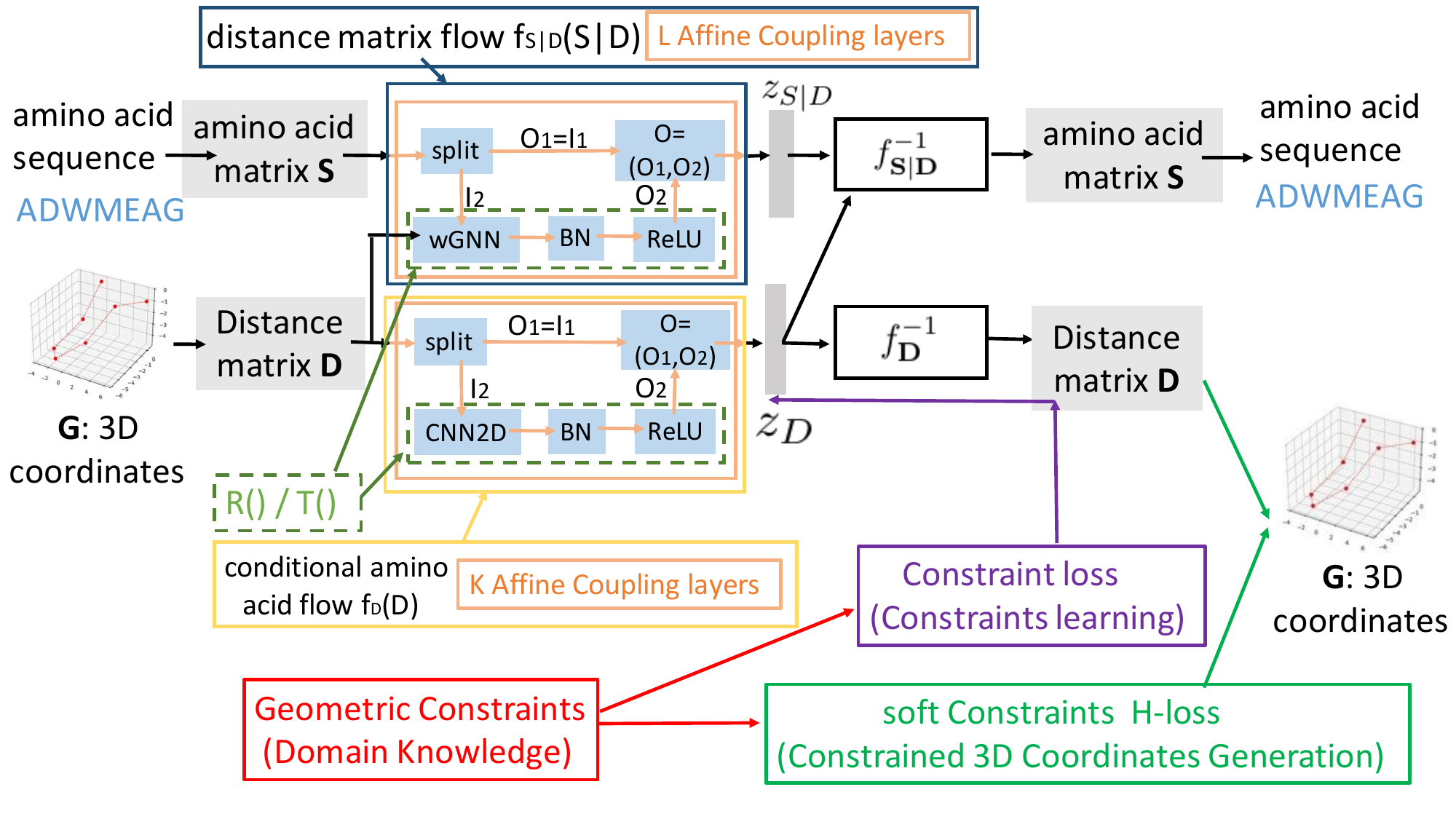}
\caption{The whole framework of \mname. 
(Forward) Flow model $f$ incorporates (i) ${\bfz}_{\bfD} = f_{\bfD}(\bfD)$ (Sec.~\ref{sec:distance}) and (ii) ${\bfz}_{\sS|\bfD} = f_{\sS|\bfD}(\sS|\bfD)$ (Sec.~\ref{sec:amino}); 
Inverse flow model $f^{-1}$ incorporates (i) $D=f_{\bfD}^{-1}({\bfz}_{\bfD})$ (Sec~\ref{sec:distance}) and (ii) $S = f_{\sS|\bfD}^{-1}({\bfz}_{\sS|\bfD}; \bfz_{\bfD})$ (Sec~\ref{sec:amino}); 
``Constraint learning'' (Sec.~\ref{sec:constraint}) minimizes constraint loss (Eq.~\ref{eqn:constraint_loss}) to encourage flow model to learn these constraints. 
``Constrained 3D coordinates generation'' (Sec~\ref{sec:coordinate_generation}) generates 3D coordinates based on distance matrix and validity constraints of CDR loops defined in Eq.~\eqref{eqn:bond_length} and~\eqref{eqn:open_loop}.   
}
\label{fig:pipeline}
\end{figure}

\section{Related Work}
\label{sec:related}

% Next, we review the related literature on (I) antibody/protein design, (II) flow models, and (III) 3D graph generation models. 

\noindent\textbf{(I) \textit{de novo} antibody/protein design} is about constructing antibodies/proteins from scratch with desirable properties. 
For protein design, variational autoencoder (VAE) based methods~\cite{sinai2017variational,costello2019hallucinate} and generative adversarial network (GAN) based methods~\cite{repecka2021expanding,karimi2020novo} were proposed to generate novel amino acid sequences. 
For antibody design, \cite{pantazes2010optcdr} combines CDR canonical structures and iteratively redesigns their positions; 
\cite{liu2020antibody} designed a neural network ensemble to backpropagate the gradient to update the amino acid sequence in continuous space; 
\cite{saka2021antibody,akbar2021silico} leverage Long Short Term Memory (LSTM) to generate amino acid sequences; 
\cite{jin2021iterative} proposed an \textit{iterative refinement graph neural network} to jointly design 3D graph structure and amino acid sequence. 
However, almost all of these methods incrementally grow the sequence by adding a single amino acid at a time. In contrast, our method generates the whole CDR loop together so that the global information (geometric constraints) is considered.

% \noindent\textbf{\textit{Inverse protein folding}}. 
% Another category of protein design method is \textit{inverse folding}. Conditioned on the 3D graph structure, a.k.a. backbone, the goal of inverse protein folding is to recover the amino acid sequence. 
% Most of the inverse folding methods are also based on deep generative models, leveraging structured transformer~\cite{ingraham2019generative}, three-dimensional convolutional neural network (3DCNN)~\cite{zhang2020prodconn,qi2020densecpd}, graph convolutionary network (GCN)~\cite{strokach2020fast}, joint sequence-folding embedding model~\cite{cao2021fold2seq}. 

\noindent\textbf{(II) Flow model} (a.k.a. normalizing flow model) is a special category of deep generative models that designs invertible mapping so that the input data can be exactly reconstructed~\cite{rezende2015variational}. 
The advantage of a flow model is its ability to reconstruct input data exactly and directly optimize the likelihood, instead of optimizing a surrogate, e.g., VAE optimizes the lower bound of likelihood~\cite{kingma2013auto}. 
It is successfully applied to image generation/impainting~\cite{horvat2021denoising,pumarola2020c}, machine translation~\cite{tang2021continuous}, and small molecule generation~\cite{shi2019graphaf,madhawa2019graphnvp,zang2020moflow,satorras2021n}. 
The conditional flow model is a well-known variant of the flow model and is also widely applied~\cite{pumarola2020c,madhawa2019graphnvp,zang2020moflow}. 
In this paper, we address some challenges of the flow model in 3D graph generation. 
Besides, \cite{pumarola2020c} proposes conditional flow on computer vision. 

\noindent\textbf{(III) 3D graph generative model} have been proposed for drug discovery recently, e.g., \cite{jin2021iterative} (mentioned in I) and \cite{luo20213d}.   
\cite{luo20213d} leveraged autoregressive 3D graph neural network to design small-molecule conformation conditioned on the pocket conformation of the target protein. 
However, these methods focus on generating a single component (e.g., an amino acid or an atom) at a time based on conditional distribution without global constraints. 
In contrast, our method generates the 3D graph at once and imposes global constraints that encode the geometric shape of CDR loops.

\begin{table}[tb]
\small 
\centering
\caption{Mathematical notations. }
\resizebox{0.74\columnwidth}{!}{
\begin{tabular}{c|l}
\toprule[1pt]
 Notations & Explanations \\ 
\midrule 
$\xx = (\bfD, \bfS)$ & 3D CDR loop \\
$N$ & number of amino acids in CDR loop \\
$V, |V|=20$ & vocabulary of amino acids \\ 
$\bfG = [\bfg_1,\cdots, \bfg_N]^{\top}$ & {3D coordinates of amino acids}, $\bfG\in\RB^{N\times 3}$ \\ 
$\sS \in \{0,1\}^{N\times |V|}$ & amino acid matrix \\
$\bfD \in \RB_{+}^{N\times N}$ & pairwise distance matrix \\ 
$f$ (i.e., $f_{\theta}$),  & flow model, learnable parameters $\theta$ \\
$\frac{\partial f_\theta}{\partial \xx}$, i.e., $\frac{\partial {\bfz}}{\partial \xx}$ & Jacobian of $f$ \\ 
$\det$ & determinant \\ 
$\eta_1, \eta_2, \epsilon_1, \epsilon_2$ & validity constraint, Eq.~\eqref{eqn:bond_length},~\eqref{eqn:open_loop} \\
$H(y; a, b, \delta)$ & constraint surrogate function, Eq.~\eqref{eqn:huber} \\
$h(\bfD)$ & constraint loss function of $\bfD$, Eq.~\eqref{eqn:constraint_loss} \\ 
$\xi_1, \xi_2, \xi_3$ / $\lambda_1, \lambda_2$ & hyperparameter in Eq.~\eqref{eqn:constraint_loss} /~\eqref{eqn:unconstrained_optimization} \\ 
\bottomrule[1pt]
\end{tabular}}
\label{table:notation}
\end{table}

\section{Proof of Corollary~\ref{cor:invariance}}
\label{sec:proof}

\noindent\textbf{Corollary 1}. 
Suppose $\bfG' \in \RB^{N\times 3}$ are the 3D coordinates of the graph by rotating and translating the original graph whose coordinates are $\bfG \in \RB^{N\times 3}$, $\bfD \in \RB^{N\times N}$ and $\bfD'\in \RB^{N\times N}$ are the distance matrix of $\bfG$ and $\bfG'$, respectively, then we have $\bfD = \bfD'$. 

\begin{proof}

\marked{Suppose $\bfD$ and $\bfD'$ are the distance matrix before and after transformation (rotation/translation), respectively. 

First, regarding rotation, without loss of generalization, we suppose the rotation matrix is $\bfQ \in \RB^{3\times 3}$, which is an orthogonal matrix satisfying $\bfQ^\top\bfQ = \bfQ\bfQ^\top = \bfI_3$, $\bfI_3\in \RB^{3\times 3}$ is identity matrix.  
Rotating 3D graph whose coordinates are $\bfG = [\bfg_1, \cdots, \bfg_N]$ results in a new 3D graph whose coordinates are $\bfG' = [\bfQ\bfg_1, \cdots, \bfQ\bfg_N]$. Then based on the definition of the distance matrix, we have 
\begin{equation}
\label{eqn:rotation}
\begin{aligned}
\bfD'_{i,j} & = ||\bfQ\bfg_i - \bfQ\bfg_i||_2 \\
& = || \bfQ (\bfg_i - \bfg_j) ||_2 \\
& = \sqrt{(\bfg_i - \bfg_j)^\top \bfQ^\top \bfQ (\bfg_i - \bfg_j)}\\
& = \sqrt{(\bfg_i - \bfg_j)^\top (\bfg_i - \bfg_j)}\\
& = ||\bfg_i - \bfg_j||_2 \\
& = \bfD_{i,j}. 
\end{aligned}
\end{equation}

Secondly, translating 3D graph (whose coordinates are $\bfG = [\bfg_1, \cdots, \bfg_N]$) by $\bfv\in \RB^{3}$ results in a new 3D graph whose coordinates are $\bfG' = [\bfg_1 + \bfv, \cdots, \bfg_N+\bfv]$. 
We have 
\begin{equation}
\label{eqn:translation}
\bfD'_{i,j} = ||(\bfg_{i} + \bfv) - (\bfg_{j} + \bfv)||_2 = ||\bfg_i - \bfg_j||_2 = \bfD_{i,j}.
\end{equation}

Third, the joint transformation of rotation and translation can be represented as 
\[
\bfG' = [\bfQ\bfg_1 + \bfv, \cdots, \bfQ\bfg_N+\bfv]. 
\]
Then based on definition of distance matrix and combining Equation~\eqref{eqn:rotation} and~\eqref{eqn:translation}, we have 
\begin{equation}
\begin{aligned}
\bfD'_{i,j} & = ||(\bfQ\bfg_i + \bfv) - (\bfQ\bfg_i + \bfv)||_2 \\
& = ||\bfQ\bfg_i - \bfQ\bfg_i||_2 \\
& = \bfD_{i,j}. 
\end{aligned}
\end{equation}

In sum, the distance matrix $\bfD$ is invariant w.r.t. translation and rotation on the 3D graph. }

\end{proof}

\section{Learning and Inference}
\label{sec:learning}

In this section, we describe the learning and inference process. 
The key steps of our method are summarized in Algorithm~\ref{alg:main}. 

During the learning process, we alternatively conduct maximum likelihood learning and constraint learning. 
Given the data sample $\xx$, maximum likelihood learning (MLE) maximizes the exact log-likelihood (Equation.~\ref{eqn:loglikelihood}) of input data w.r.t. parameters $\theta$, 
\begin{equation}
\label{eqn:mle_obj}
\begin{aligned}
& \calL_{\text{MLE}}(\theta) =  \log p_{{\xx}}({\xx}) \\
= & \log p_{\bfz}({\bfz}) + \log \big(\big|\det \frac{\partial {\bfz}}{\partial {\xx}} \big| \big) \\  
 = & \log p_{{\bfz}_{\bfD}}({\bfz}_{\bfD}) + \log p_{{\bfz}_{\sS|\bfD}}({\bfz}_{\sS|\bfD}) \\
 & + \log \bigg( \bigg| \det \frac{\partial f_{\bfD}}{\partial \bfD} \bigg| \bigg)  + \log \bigg( \bigg| \det \frac{\partial f_{\sS|\bfD}}{\partial \sS} \bigg| \bigg), 
\end{aligned}
\end{equation}
where latent variable $\bfz$ satisfies $\bfz = f_{\theta}(\xx)$, 
the first two term $p_{{\bfz}_{\bfD}}(\cdot)$, $p_{{\bfz}_{\sS|\bfD}}(\cdot)$ follow isotropic Gaussian distribution, the third term $\log \big( \big| \det \frac{\partial f_{\bfD}}{\partial \bfD} \big| \big)$ is given in Section~\ref{sec:distance}; the fourth term $\log \big( \big| \det \frac{\partial f_{\sS|\bfD}}{\partial \bfS} \big| \big)$ is defined in Section~\ref{sec:amino}.

During inference process, we sample latent variable ${\bfz} = ({\bfz}_{\bfD}, {\bfz}_{\sS|\bfD})$ from isotropic Gaussian distribution, generate the distance matrix $\bfD$, and transform $\bfD$ into 3D coordinates $\bfG$ following~Section~\ref{sec:distance}. 
Then we generate $\sS$ conditioned on $\bfD$. 
For each row in $\sS\in \RB^{N\times |V|}$, we use argmax operation to the $|V|$-dimensional row vector to obtain the corresponding amino acid's category. 
The whole pipeline is illustrated in Figure~\ref{fig:pipeline}.

\section{Data and Preprocessing}
\label{sec:data}
This section describes the datasets and the preprocessing procedure. 

% \begin{itemize}[leftmargin=*]
\noindent (1) \textbf{SabDab} (the structural antibody database)\footnote{It is publicly available \url{http://opig.stats.ox.ac.uk/webapps/newsabdab/sabdab/}. } is a database containing all the antibody structures available in the PDB (Protein Data Bank), annotated and presented in a consistent fashion~\cite{dunbar2014sabdab}. 
It has 5,494 antibodies. One antibody may contains multiple CDR loops (both amino acid sequence and 3D structure). 
After filtering the antibodies whose CDR structures are not available, we get 8,381 data points. 
For H1 loops, the average length of CDR loops (i.e., number of amino acids) is 7.2, the standard deviation is 0.6; 
for H2 loops, the average length is 5.9, with a 0.7 standard deviation; for H3 loops, the average is 12.3, and the standard deviation is 4.4. 
The data points are randomly split into training, validation, and test sets with a ratio of 8:1:1. The data is saved in the format of a PDB (Protein Data Bank) file, where the coordinates of all the atoms are available. 
We use the coordinate of $\alpha$-carbon atom to denote the coordinate of amino acid, following~\cite{ingraham2019generative,jin2021iterative}. 
The unit of distance in PDB file is $\mathring{A}$, $1\mathring{A}=10^{-10}m = 0.1nm$. 

\noindent (2) \textbf{CoV-AbDab} (the coronavirus antibody database)~\cite{raybould2021cov}\footnote{It is publicly available at \url{http://opig.stats.ox.ac.uk/webapps/covabdab/}} documents all published/patented binding antibodies and nanobodies to coronaviruses, including SARS-CoV2, SARS-CoV1, and MERS-CoV. 
It contains 2,942 antibodies and 494 nanobodies. The 3D structures are not available. 
In this paper, we restrict our attention on CDR H3 sequence of antibodies, and ignore nanobodies. 
We use half of the data (1.5K) to train machine learning model to evaluate SARS-CoV-2 neutralization ability (as mentioned in Section~\ref{sec:metric}), it is a binary classification task, the train/validation/test split ratio is 8:1:1. 
The remaining data are used to train H3 loops generation models, where train/validation split ratio is also 9:1. 
Test data is not required because the generated H3 loops are evaluated using the trained model that evaluates SARS-CoV-2 neutralization ability. 
% \end{itemize}

\section{Evaluation Metrics}
\label{sec:metric}

In this section, we describe the metrics for evaluating the performance of all the methods. 
% We have two major metrics: (i) neutralization oracle that measures a CDR H3 loop's neutralization ability to SARS-CoV virus; and (ii) protein perplexity that measures the protein-likeness of a CDR H3 loop. 

\begin{itemize}[leftmargin=*]
\item 
\noindent (1) \textbf{Protein Perplexity (PPL)}. 
Protein perplexity is the exponential of the average log-likelihood and measures how well a probability model predicts amino acid sequence~\cite{ingraham2019generative,armenteros2020language}. We train an LSTM model as language model. Lower perplexity indicates better performance. 

\item 
\noindent (2) \textbf{Root-Mean-Square Deviation (RMSD)} measures the alignment between tested conformations $R\in \mathbb{R}^{N\times 3}$ and reference conformation ${R}^{r}\in \mathbb{R}^{N\times 3}$, $N$ is the number of points in the conformation, defined as 
\[\text{RMSD}(R,\hat{R}) = \big(\frac{1}{N} \sum_{i=1}^{N} ||R_i - \hat{R}_i||_2^2 \big)^{\frac{1}{2}},\] 
where $R_i$ is the $i$-th row of conformation $R$. 
The conformation $\hat{R}$ is obtained by an alignment function $\hat{R} = A(R, R^{r})$, which rotates and translates the reference conformation $R^r$ to have the smallest distance to the generated R according to the RMSD metrics, which is calculated by the Kabsch algorithm~\cite{kabsch1976solution}. 
Lower RMSD indicates better performance.

% Two common metrics to evaluate the quality of the generated conformations are \textbf{Coverage} (COV) and \textbf{Matching} (MAT) scores. 
% \begin{equation}
% \text{COV}(S_g, S_r) = \frac{1}{|S_r|}|\{R \in S_r|\text{RMSD}(R,\hat{R})<\delta, \hat{R} \in S_g\}|,
% \end{equation}
% \begin{equation}
% \text{MAT}(S_g, S_r) = \frac{1}{|S_r|}\sum_{R \in S_r} \min_{\hat{R} \in S_g} \text{RMSD}(R, \hat{R}),
% \end{equation}
% where $S_g$ and $S_r$ denote generated and reference conformations, respectively. $\delta$ is a given \textbf{RMSD} threshold. 
\item 
\noindent (3) \textbf{Validity Rate (VR)} is the percentage of valid CDR loops in all the generated CDRs. A CDR loop is valid if it satisfies two criteria in Equation~\eqref{eqn:bond_length} and~\eqref{eqn:open_loop}. 
Higher validity is more desirable. 

\item 
\noindent (4) \textbf{Similarity} is evaluated on amino acid sequence level. 
The similarity between two sequences $s_1$ and $s_2$, and is defined as 
\begin{equation}
\label{eqn:similarity}
\text{sim}(s_1,s_2) = \frac{|\text{LCS}(s_1, s_2)|}{\max\{|s_1|, |s_2|\}},
\end{equation}
where $|s|$ denotes the length of sequence $s$, $\text{LCS}(s_1, s_2)$ denotes the longest common substring between $s_1$ and $s_2$, computed by dynamic programming. 
The similarity value ranges from 0 to 1, a higher value indicates more similarity. 
The distance is defined as one minus similarity. 
The similarity is not used directly in our experiment. 
It is used for evaluating diversity. 

\item 
\noindent (5) \textbf{Diversity (Div)} is defined as the average pairwise distance between the CDR loops. 
Specifically, suppose $\mathcal{S}$ is the set of all the generated CDR loops, we have 
% \begin{equation*}
\[
\text{diversity}(\mathcal{S}) = 1 - \frac{1}{|\mathcal{S}|(|\mathcal{S}|-1)}\sum_{s_1,s_2 \in \mathcal{S}, s_1 \neq s_2} \text{sim}(s_1, s_2),
\]
% \end{equation*}
where $\text{sim}(\cdot,\cdot)$ is the similarity function defined above. 
Diversity ranges from 0 to 1, higher diversity is more desirable.

\item 
\noindent (6) \textbf{SARS-CoV Neutralization Ability}. 
Following~\cite{jin2021iterative}, we train a machine learning model to predict the neutralization ability of a CDR-H3 sequence to either SARS-CoV-1 or SARS-CoV-2 viruses. 
The predictor takes as CDR-H3 sequence of an antibody and outputs a neutralization probability for the SARS-CoV-1 and SARS-CoV-2 viruses. 
Each residue is embedded into a 100-dimensional vector. 
Then the embeddings are fed to a three-layer bi-LSTM (bidirectional Long Short Term Memory)~\cite{huang2015bidirectional}, followed by average-pooling and a two-layer MLP (multi-layer perceptron). 
The hidden dimension of bi-LSTM is 100. 
The activate functions in the hidden layers and output layer are ReLU and sigmoid, respectively. 
The prediction is very accurate, achieving over 0.99 ROC-AUC and over 0.95 F1 score. 
The neutralization predictor yields neutralization probability (ranging from 0 to 1), higher probability is desirable. 
\end{itemize}

\section{Baseline Methods}
\label{sec:baseline}

In this section, we describe the baseline methods and briefly introduce the implementation details. 
All these baselines are state-of-the-art approaches in antibody design. 
We use the default setup following the original paper. 

\begin{enumerate}[leftmargin=*]
\item
\noindent\textbf{Reference} evaluates the metrics of real CDR loops in the test set. 

\item 
\noindent\textbf{LSTM}~\cite{saka2021antibody,akbar2021silico} leverages LSTM (long short-term memory) to generate amino acid sequence and does not model 3D structure. 
Following the original paper, it is trained by the Adam optimizer and has a learning rate of 0.01. Dropout rates were chosen from 0.1 and 0.2 to regularize all layers
The hidden dimension of LSTM is set to 256. 
% loop in autoregressive manner.  
% \item \textbf{Ens-Grad} (Ensemble Gradient)~\cite{liu2020antibody} designed a neural network ensemble to back-propagate the gradient to update the amino acid sequence in continuous space. enrichment. 
\item 
\noindent\textbf{GA (genetic algorithm)} starts from a population of CDR loops. Given the data in the current population, to generate offspring for the next generation, we leverage the following two operations: (i) crossover (switching subsequence between two loops) and (ii) mutation (randomly changing one amino acid) to produce new candidate in each iteration~\cite{liu2017computational}. The population size of GA is set to 100, and the number of generations is set to 1K, which is sufficient for GA to converge. In each generation, when producing the offspring, we generate 1K offsprings, where half of them are produced through crossover operation, and half are produced through mutation operation. We use perplexity as the oracle to select the most promising offspring for the next generation. 

\item 
\noindent\textbf{IR-GNN} (Iterative refinement graph neural network)~\cite{jin2021iterative} first unravels amino acid sequence and iteratively refine its predicted global structure. The inferred structure in turn guides subsequent amino acid choices. Following their original paper, for the GNN, both its structure and sequence MPN have four message passing layers, with a hidden dimension of 256 and block size b = 4. It is trained by the Adam optimizer with a dropout of 0.1 and a learning rate of 0.001. 

\item 
\noindent\textbf{AR-GNN} (auto-regressive graph neural network)~\cite{you2018graphrnn,luo20213d} was proposed for molecule generation and was adapted for antibody CDR design~\cite{jin2021iterative}. 
The neural architecture of AR-GNN follows the setups of IR-GNN. 
Different from IR-GNN that uses iterative refinement mechanism, AR-GNN generates the amino acid sequence and 3D structural graph autoregressively. 

\item 
\noindent\textbf{\mname \ - constraint} is the variant of \mname that does not consider validity constraint (Section~\ref{sec:constraint} and~\ref{sec:coordinate_generation}). The other setups are the same as \mname (Section~\ref{sec:implement}). 
\end{enumerate}

LSTM, IR-GNN and AR-GNN fall into the category of deep generative models while GA are belonging to traditional heuristic searching method. 
\cite{satorras2021n} is an ODE type of flow model. However, the training process is computationally expensive since the same forward operation has to be done multiple times sequentially in order to solve the ODE equation. It also exhibited some instabilities~\cite{satorras2021n} and fails in our experiment, so it is not included as baseline.  
All the approaches exhibit state-of-the-art performance on antibody design~\cite{saka2021antibody,akbar2021silico,jin2021iterative}.

\section{Implementation Details}
\label{sec:implement}

In this section, we elaborate on the implementation details for future reproduction, including network architecture, hyperparameter setting, etc. 
It is implemented using Pytorch 1.7.1, and Python 3.7.1. 
The model is trained on NVIDIA Pascal Titan X GPU. 
We also upload the code repository in the supplementary materials. 

We use 5 affine coupling layers for the distance matrix flow model and 10 affine coupling layers for the conditional amino acid flow model. We adopt two 3 $\times$ 3 convolution layers with 128 dimensions in the hidden state of every coupling layer. For each graph coupling layer, we set one wGNN layer (Sec~\ref{sec:amino}) with 64 dimensions followed by a two-layer multilayer perceptron with 128 \& 64 hidden dimensions. 
$\xi$ in Equation~\eqref{eqn:wgnn} is set to 0.3. 
The model is trained in an end-to-end manner, where we use Adam optimizer with initial learning rate $1e^{-3}$, the batch size is set to 64, and the number of training epochs is set to 200. 
The model is trained on a NVIDIA Pascal Titan X GPU, the training process takes around 3 hours for each of the three tasks on SAbDab. 

In constraint learning (Section~\ref{sec:constraint}), the hyperparameters $\xi_1, \xi_2, \xi_3$ in Equation~\eqref{eqn:constraint_loss} are set to $\xi_1 = 10, \xi_2 = 50, \xi_3 = 1$. 
We alternatively optimize log-likelihood in Equation.~\eqref{eqn:loglikelihood} and constraint loss in Equation.~\eqref{eqn:constraint_loss}. 
Constraint learning is only conducted during the learning procedure and is not utilized in inference.

\begin{figure}
\centering
\includegraphics[width=0.5\columnwidth]{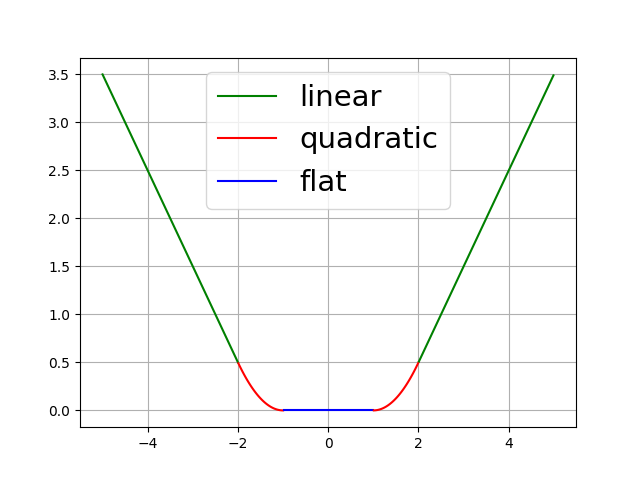}
\caption{An illustration of H function (Eq.~\ref{eqn:huber}), where $a$=$-1$, $b$=$1$, $\delta$=$1$. It has equal values and slopes of the different sections at the four connection points ($-2, -1, 1, 2$) so it is differentiable everywhere. }
\label{fig:H}
\end{figure}

\begin{figure}
\centering
\includegraphics[width=0.86\columnwidth]{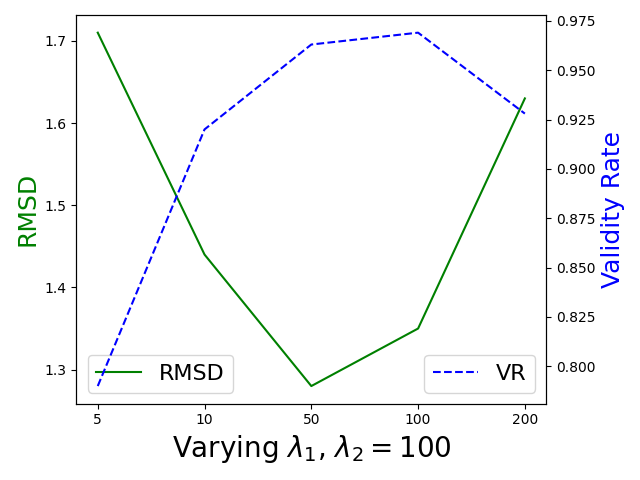}
\includegraphics[width=0.87\columnwidth]{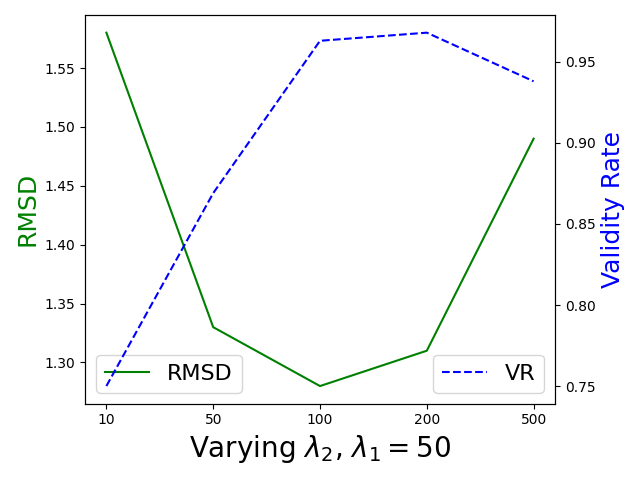}
\caption{Sensitivity Analysis on $\lambda_1$ and $\lambda_2$. We find the combination $\lambda_1=50$ and $\lambda_2=100$ works best empirically. 
}
\label{fig:sensitive_lambda}
\end{figure}

In 3D coordinates generation (Equation.~\ref{eqn:unconstrained_optimization} in Section~\ref{sec:coordinate_generation}), to make sure that the constraints are satisfied, we set $\lambda_1=50$ and $\lambda_2=100$. 
We conduct sensitivity analysis about $\lambda_1$ and $\lambda_2$ in Section~\ref{sec:sensitive} to justify their choices. 
Hyperparameters $\eta_3$ and $\epsilon_3$ are set to $\eta_3 = \eta_2 - \eta_1$, $\epsilon_3 = \epsilon_2 - \epsilon_1$. 
We use SGD (stochastic gradient descent) as optimizer when optimizing Equation~\eqref{eqn:unconstrained_optimization}. 
Worth to mention that the constrained 3D coordinates generation is not conducted during the learning procedure and is only conducted during inference.

The setup of $\eta_1,\eta_2, \epsilon_1, \epsilon_2$ is based on domain knowledge~\cite{stanfield2014antibody,nowak2016length} and empirical validation. 
Concretely, for H1 loop, $\eta_1=3.76$, $\eta_2=3.84$, $\epsilon_1=11.4, \epsilon_2 = 13.1$; for H2 loop, $\eta_1=3.76$, $\eta_2=3.87$, $\epsilon_1=5.0, \epsilon_2 = 5.9$; for H3 loop, $\eta_1=3.71$, $\eta_2=3.88$, $\epsilon_1=6.5, \epsilon_2 = 8.5$. 
The units of $\epsilon_1, \epsilon_2, \eta_1, \eta_2$ are Angstrom $\mathring{A}$ ($10^{-10}$ m, i.e., 0.1 nm). 
We find that the bond lengths between amino acids are relatively fixed across H1, H2, and H3 loops, whereas the lengths of open loops vary greatly. 
The hyperparameters $\epsilon_3, \eta_3$ are set to $\epsilon_3 = \epsilon_2 - \epsilon_1$ and $\eta_3 = \eta_2 - \eta_1$. 

On each generation task (H1, H2, H3), we generate 10K CDR loops and evaluate their VR (validity rate), PPL (perplexity), and Div (Diversity). 
When evaluating geometric graph metrics RMSD, for each tested CDR loop in the test set, we select the CDR loop that has the smallest RMSD with the test CDR loops from the set of all generated CDR loops and evaluate the RMSD. 

\begin{figure}
\centering
\includegraphics[width=0.86\columnwidth]{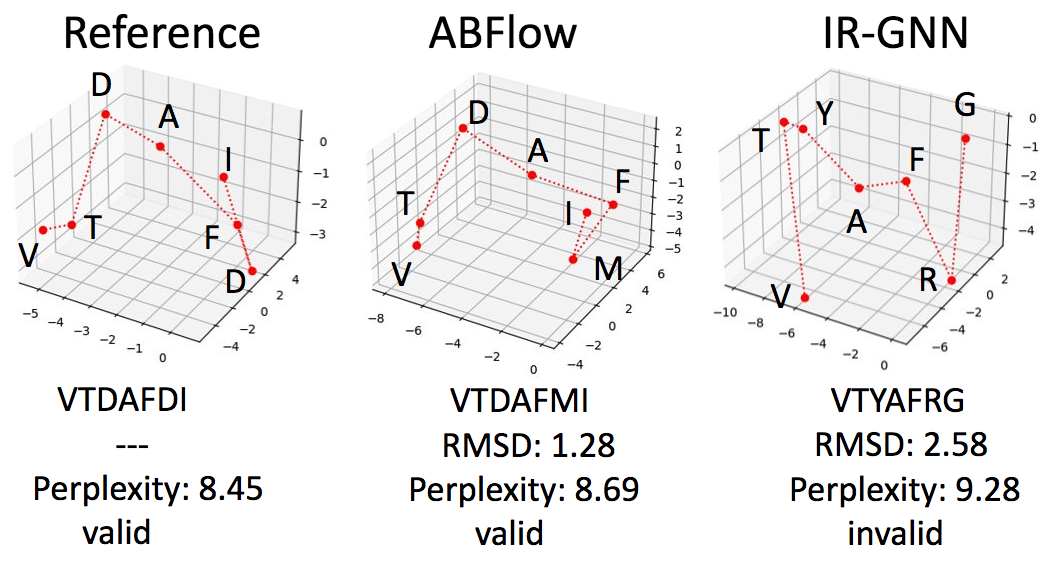}
\caption{Case study of CDR H3 structure of \textbf{Reference}, \textbf{\mname}, \textbf{IR-GNN} (the best baseline). The H3 loop is from the antibody whose PDB ID is ``3s34''. We found that (i) on amino acid sequence level, the similarity (defined in Equation~\ref{eqn:similarity}) between ``VTDAFMI'' (generated by \mname) and ``VTDAFDI'' (the reference) is 0.86 ($\frac{6}{7}$), whereas the similarity between ``VTYAFRG'' (IR-GNN) and ``VTDAFDI'' (the reference) is 0.57. 
Also, \mname got lower RMSD and perplexity than IR-GNN.  
}
\label{fig:casestudy}
\end{figure}

% As for the optimization experiments, we further train a regression model to map the latent embeddings to different property scalars (discussed in Sec. 5.3 and 5.4) by a multi-layer perceptron with 18-dim linear layer -> ReLu -> 1-dim linear layer structures. 

\section{Sensitivity Analysis}
\label{sec:sensitive}

This section conducts sensitive analysis on some critical hyperparameters. 
In 3D coordinates generation (Equation.~\ref{eqn:unconstrained_optimization} in Section~\ref{sec:coordinate_generation}), we use $\lambda_1$ and $\lambda_2$ to incorporate soft constraints to make sure that the hard constraints (defined in Equation~\ref{eqn:bond_length} and~\ref{eqn:open_loop}) are satisfied. 

We conduct sensitive analysis about $\lambda_1$ and $\lambda_2$ on cdr H1 loop design tasks. 
Specifically, regarding $\lambda_2$ $\lambda_2$, we fix one and vary the other using grid search, then record the results in terms of RMSD (root mean square deviation) and validity rate. 
In Figure~\ref{fig:sensitive_lambda}, we plot the RMSD and validity as a function of $\lambda_1$ (fixing $\lambda_2$) and $\lambda_2$ (fixing $\lambda_1$), respectively. 
We find the combination $\lambda_1=50$ and $\lambda_2=100$ works best empirically. 
We apply the similar strategy to select other hyperparameters, e.g., $\xi_1, \xi_2, \xi_3$ in Equation~\ref{eqn:constraint_loss}.

\section{Case Study}
\label{sec:casestudy}

This section shows a case study in Figure~\ref{fig:casestudy} and analyzes the results. 
Specifically, 
we show CDR H3 structure of {Reference}, {\mname}, {IR-GNN} (the best baseline). The H3 loop is from the antibody whose PDB ID is ``3s34''. We found that (i) on amino acid sequence level, the similarity (defined in Equation~\ref{eqn:similarity}) between ``VTDAFMI'' (generated by \mname) and ``VTDAFDI'' (the reference) is 0.86 ($\frac{6}{7}$), whereas the similarity between ``VTYAFRG'' (IR-GNN) and ``VTDAFDI'' (the reference) is 0.57. 
Also, \mname got lower RMSD and perplexity than IR-GNN.  
The case study validates \mname's superiority.

\end{document}